\documentclass[runningheads]{llncs}

\usepackage[mobile]{eccv}

\usepackage{eccvabbrv}
\usepackage{graphicx}
\usepackage{amsmath,amssymb,amsfonts}
\usepackage{booktabs}
\usepackage{multirow}
\usepackage{colortbl}
\usepackage{xcolor}
\usepackage{float}
\usepackage[accsupp]{axessibility}
\usepackage[breaklinks,colorlinks,citecolor=eccvblue]{hyperref}

\newcommand{\repourl}{https://github.com/bibeshpyakurel/SLAPBench}
\definecolor{aucPerfect}{HTML}{3DBC6E}
\definecolor{aucStrong}{HTML}{8FCB7F}
\definecolor{aucMid}{HTML}{D7DC79}
\definecolor{aucWeak}{HTML}{F0A85C}
\definecolor{aucBad}{HTML}{E8736A}
\definecolor{collapseRed}{HTML}{F7D4D7}
\definecolor{funcGreen}{HTML}{D6EFD9}
\definecolor{grpBin}{HTML}{FBEFD9}
\definecolor{grpScore}{HTML}{E2ECFB}

\graphicspath{{figures/}}

\begin{document}

\title{SLAPBench: Benchmarking Multimodal Large Language Models
       for Four-Finger SLAP Fingerprint Verification}
\titlerunning{SLAPBench: MLLMs for Four-Finger SLAP Verification}

% TODO (optional): add ORCIDs.
\author{Bibesh Pyakurel \and M.~G.~Sarwar Murshed}
\authorrunning{B.~Pyakurel and M.~G.~S.~Murshed}
\institute{Computer Science Department,
           University of Wisconsin--Green Bay, Green Bay, WI, USA}

\maketitle

\begin{abstract}
Four-finger SLAP fingerprints are flat live-scan impressions of the index,
middle, ring, and little fingers of one hand. They are widely used for identity
verification in high-stakes settings, including border control, immigration
screening, law enforcement, criminal background checks, and national identity
management. Despite this operational importance, no benchmark has evaluated
whether multimodal large language models (MLLMs) can verify identity from SLAP
fingerprint images. We introduce \textbf{SLAPBench}, the first benchmark for
MLLM-based four-finger SLAP fingerprint verification, built from NIST Special
Database 302b (SD302b), an operational nail-to-nail fingerprint dataset
containing 201 participants captured at 500 and 1000~PPI across multiple
devices. We construct 7,832 verification pairs using an impostor-heavy
all-pairs protocol, comprising 176~mated and 7,656~non-mated pairs.
We evaluate four open-source MLLMs, InternVL3-8B-Instruct,
Qwen2.5-VL-7B-Instruct, Qwen3-VL-8B-Instruct, and Gemma-3-12B-IT, alongside the
proprietary Claude Opus~4.8 model under three prompting strategies: zero-shot,
task-description, and continuous similarity scoring. Prompt design has a
substantial effect on verification behavior. Task-description prompting causes
all open-source models to collapse to near-100\% False Accept Rate~(FAR), while
Gemma-3-12B-IT also collapses under zero-shot prompting. Claude Opus~4.8 is the
only model that avoids collapse under both binary prompts and achieves the best
binary result, with FAR~$= 20.2\%$.
Continuous similarity scoring removes collapse across the open-source models
and exposes clear differences in discrimination. Claude Opus~4.8 achieves
AUC~$= 0.953$ and EER~$= 11.8\%$, followed by Gemma-3-12B-IT with
AUC~$= 0.837$ and EER~$= 15.1\%$. InternVL3-8B-Instruct shows inverted
calibration (AUC~$= 0.590$), and Qwen2.5-VL-7B-Instruct remains near random
(AUC~$= 0.567$). Qwen3-VL-8B-Instruct obtains perfect separation
(AUC~$= 1.000$, EER~$= 0.0\%$); however, we treat this as a diagnostic rather
than as evidence of biometric capability. Because SD302b provides only one SLAP
capture per finger position, mated pairs must be formed across resolutions,
raising the concern that a model could exploit resolution rather than identity.
We run a matched-resolution control that makes both classes cross-resolution,
and the perfect separation persists (AUC~$= 1.000$), which rules the resolution
shortcut out. What remains, and cannot be tested within SD302b, is that a mated
pair is one capture rendered twice, so the model may instead be detecting
near-duplicate images. We therefore read this as a cautionary finding about how
MLLM biometric benchmarks are assembled. Finally, a stratified analysis over
gender, race, and age suggests that demographic disparity increases when model
discrimination is weak, although subgroup sizes limit this to an initial
fairness probe. Overall, SLAPBench establishes the first SLAP-specific MLLM
verification baseline and shows that prompting governs collapse, while model
capability governs whether meaningful fingerprint discrimination can be
recovered.
\end{abstract}

% ────────────────────────────────────────────────────────────────────────────
\section{Introduction}
\label{sec:intro}

Four-finger SLAP fingerprints are flat live-scan impressions acquired by
simultaneously capturing the index, middle, ring, and little fingers of one
hand. These impressions are widely used for identity verification in
high-stakes operational settings, including border control, immigration
screening, law enforcement, criminal background checks, and national identity
management. In U.S. border operations, SLAP impressions support traveler
identity verification against large-scale biometric repositories such as the
DHS Automated Biometric Identification System (IDENT)~\cite{nist_sd302}.

Despite their operational importance, SLAP fingerprints remain largely
unexplored in recent evaluations of multimodal large language models (MLLMs).
Existing fingerprint-oriented MLLM benchmarks primarily evaluate individual
rolled or plain fingerprint impressions~\cite{fpbench2025}. For example,
FPBench evaluates 20~MLLMs across eight fingerprint tasks, but its evaluations
are based on single-finger rolled or plain impressions drawn from Fingerprint
Verification Competition (FVC) datasets and NIST
SD302d/SD301a~\cite{fpbench2025}. These settings differ substantially from SLAP
verification. A four-finger SLAP image contains multiple fingers in a single
frame and exhibits variation in ridge clarity, contact pressure, finger
placement, inter-finger spacing, and local distortion. Consequently, SLAP
verification requires not only fine-grained ridge comparison but also spatial
reasoning over the relative structure and correspondence of multiple fingers.

NIST Special Database 302b (SD302b) provides a particularly relevant basis for
studying this problem. Unlike curated benchmarks designed primarily for
controlled algorithmic evaluation, SD302b reflects operational live-scan
acquisition and includes images captured across multiple devices and
resolutions. This makes it a challenging and deployment-relevant testbed for
evaluating whether general-purpose MLLMs can perform identity verification from
four-finger SLAP fingerprint images.

A second open question concerns the role of prompting. Prior biometric MLLM
benchmarks for face and fingerprint recognition have reported
\emph{positive-bias collapse}, in which a model repeatedly predicts ``same
person'' regardless of the input pair~\cite{facerecbench2025,fpbench2025}. On a
balanced verification set, such behavior can yield 50\% accuracy while
concealing a critical failure mode. In high-stakes biometric applications,
however, this behavior would be unacceptable because it corresponds to
accepting every comparison as genuine. Binary forced-choice prompting is
commonly used in biometric MLLM evaluation, but it remains unclear whether this
format induces similar collapse on SLAP fingerprint images, or whether
alternative prompt designs can mitigate the problem.

To address these gaps, we introduce SLAPBench, an evaluation benchmark for
MLLM-based four-finger SLAP fingerprint verification. SLAPBench is intended as
an evaluation probe of current MLLM behavior, not as a proposed deployment
system for operational biometric verification. Our goal is to characterize how
these models succeed and fail on SLAP-format fingerprint images, and to examine
how prompting strategy affects verification behavior, collapse, and
discriminative performance.

This paper makes the following contributions:

\begin{enumerate}
  \item \textbf{SLAPBench}, the first benchmark for MLLM evaluation on
        four-finger SLAP fingerprint verification, built on NIST SD302b, an
        operational dataset that no prior fingerprint MLLM work has used.

  \item \textbf{A reproducible evaluation protocol} of 7,832~fixed pairs
        (176~mated, formed across resolution because SD302b provides a
        single SLAP capture per finger position; 7,656~non-mated, covering
        all $\binom{88}{2}$ subject combinations per FRGP), released as a
        fixed pair manifest with demographic metadata that supports a
        stratified analysis across gender, race, and age.

  \item \textbf{Systematic evidence of positive-bias collapse on SLAP data.}
        Five of ten binary prompting configurations collapse to accepting
        nearly every pair as genuine. All four open-source models collapse
        under task-description prompting and Gemma-3-12B collapses under
        zero-shot as well, while the proprietary Claude Opus~4.8 resists
        collapse under both prompts. Collapse is therefore largely a
        property of the prompt format, but a sufficiently capable model can
        overcome it.

  \item \textbf{A similarity-scoring prompt that exposes model capability.}
        Scoring eliminates collapse in all four open-source models, yet
        discrimination still ranges from functional (Claude Opus~4.8,
        Gemma-3-12B) to near-random or inverted (Qwen2.5-VL-7B,
        InternVL3-8B). Scoring is thus necessary but not sufficient, and
        model architecture sets the ceiling
        (Table~\ref{tab:main_results}).

  \item \textbf{A cautionary analysis of a perfect benchmark score.}
        Qwen3-VL-8B attains AUC~$= 1.000$ under scoring. A matched-resolution
        control that removes resolution as a class cue leaves the perfect
        separation intact, which excludes the resolution shortcut; the
        single-capture design of SD302b nonetheless leaves near-duplicate
        detection in place, and no configuration of the dataset can separate
        it from identity matching. We offer this as evidence that near-perfect
        MLLM biometric results warrant a protocol audit before they are read
        as capability.

  \item \textbf{Public release} of the evaluation code, the fixed pair
        manifest, the raw result CSVs, and the per-pair score files, so that
        every number reported here can be recomputed. Code and data:
        \url{\repourl}.
\end{enumerate}

% ────────────────────────────────────────────────────────────────────────────
\section{Related Work}
\label{sec:related}

\textbf{Fingerprint recognition.}
Classical pipelines of acquisition, enhancement, feature extraction, and
matching~\cite{maltoni_handbook} have been advanced at each stage by deep
learning: FingerNet unifies ridge-orientation and minutiae
extraction~\cite{fingerNet}, DeepPrint learns fixed-length embeddings for
scalable verification~\cite{deepprint}, and commercial matchers such as
VeriFinger~\cite{verifinger} provide operational FAR/FRR baselines. SLAP
segmentation is studied separately~\cite{murshed2024SlapSeg}; to our knowledge
no prior work evaluates end-to-end MLLM performance on SLAP images.

\textbf{MLLM biometric benchmarks.}
FaceXBench~\cite{facexbench2024} evaluates 28~models on 5{,}000 face-task
questions and finds chain-of-thought prompting consistently \emph{degrades}
performance, a caution against assuming more context helps.
FaceRecBench~\cite{facerecbench2025} benchmarks 27~MLLMs on face verification
and reports two findings we revisit for fingerprints: many models collapse to
50\% accuracy by always accepting or rejecting, and the best model carries the
strongest demographic bias; a follow-up extends this to heterogeneous face
recognition with similar capability gaps~\cite{shahreza2026hfr}.
FPBench~\cite{fpbench2025} is closest to us, evaluating 20~MLLMs on eight
single-finger fingerprint tasks under MCQ prompting. We differ in three ways:
four-finger SLAP images requiring spatial reasoning; continuous similarity
scoring alongside binary classification; and the operational SD302b dataset.
Across these benchmarks, response collapse recurs and has been attributed to
training-data priors favouring ``same'' together with the ambiguity of
biometric comparison in language; we give the first quantitative account on
SLAP images and show that continuous scoring removes it.

% ────────────────────────────────────────────────────────────────────────────
\section{Dataset and Data Preparation}
\label{sec:dataset}

\subsection{NIST Special Database 302b}

NIST SD302b~\cite{nist_sd302} is a subset of the Nail-to-Nail (N2N)
Fingerprint Challenge dataset collected by the Intelligence Advanced
Research Projects Activity (IARPA).
The database contains fingerprint images from 201~participants captured
using multiple devices at two resolutions: 500~PPI (standard operational
resolution) and 1000~PPI (high resolution for research).
Participants span diverse demographics including age (range: 18--70),
gender, race, and occupational background (work type), enabling
stratified fairness analysis.
We focus exclusively on \textbf{FRGP 13} (right four-finger SLAP) and
\textbf{FRGP 14} (left four-finger SLAP) images captured by
\textbf{Device~R}.
Device~R images in SD302b are captured natively at 1000~PPI, and the
corresponding 500~PPI versions are supplied for compatibility with
500~PPI fingerprint algorithms. The 500~PPI image is therefore a
downsampled version of the same capture rather than a second, independent
impression.
We confirmed this directly against the distributed file structure: each
(subject,~FRGP) combination is represented by exactly one capture, available at
two resolutions, and SD302b offers no repeat SLAP impression for any subject and
finger position. This database property, not a choice of ours, fixes the form of
a mated pair: a \emph{same-capture cross-resolution} comparison between the
native 1000~PPI image and its 500~PPI counterpart, not two independent
impressions (Sections~\ref{sec:pairs} and~\ref{sec:discussion}).

\textbf{Data preparation.}
Following the SD302b documentation we exclude errata-flagged subjects
(labeling errors), roll images (devices U, V), FRGP~15 thumb-slaps, and
pre-cropped segmented crops, and keep only subjects with both R-500 and R-1000
captures. Of the \textbf{201} participants, \textbf{92} have Device~R
four-finger SLAP at both FRGP~13/14 and both resolutions; removing
errata/incomplete subjects leaves \textbf{88} clean subjects. The verification
experiment therefore uses \textbf{352 Device~R images}
(88~subjects $\times$ 2~FRGP $\times$ 2~resolutions).
For each image we record image-level metadata, per-finger segmentation ground
truth (bounding boxes and rotation angle $\theta$ from the SD302b segmentation
CSVs), and participant demographics.

\textbf{Ground-truth segmentation.}
SD302b ships per-image segmentation CSVs giving bounding-box coordinates and a
rotation angle for each finger (dedicated SLAP segmentation systems address the
same step~\cite{murshed2024SlapSeg}); a representative overlaid image appears in
Fig.~\ref{fig:supp_anatomy}. For verification we use the full SLAP
images without cropping, as operational pipelines do. Because a single frame
holds four overlapping fingers of similar morphology, any two SLAP images look
globally alike whether mated or not, which underlies the positive-bias collapse
analyzed in Section~\ref{sec:discussion}.

% ────────────────────────────────────────────────────────────────────────────
\section{Evaluation Protocol and Methodology}
\label{sec:method}

\subsection{Pair Construction}
\label{sec:pairs}

We construct mated and non-mated verification pairs following the
all-pairs protocol used in FaceRecBench~\cite{facerecbench2025} and
FPBench~\cite{fpbench2025}.

\textbf{Mated (genuine) pairs.}
For each FRGP $\in \{13, 14\}$, we pair each subject's native 1000~PPI
image with the corresponding downsampled 500~PPI version of the
\emph{same capture}.
With 88~clean subjects, this yields 88~mated pairs per FRGP and 176~mated
pairs in total, fixed by the structure of the database rather than by a
sampling decision.

\textbf{Non-mated (impostor) pairs.}
For each FRGP, we form all pairings of distinct subjects.
With 88 subjects, this yields $\binom{88}{2} = 3{,}828$ unique non-mated
pairs per FRGP, and $3{,}828 \times 2 = 7{,}656$ non-mated pairs in total.

The resulting \textbf{7,832~pairs} (176~mated $+$ 7,656~non-mated), released as
a fixed manifest, form an intentionally impostor-heavy protocol; as is standard
in biometrics, we emphasize FAR, FRR, AUC, EER, and TAR at low FAR rather than
overall accuracy, which the non-mated majority dominates.

\textbf{Resolution structure and its consequence.}
Mated pairs are cross-resolution (1000~PPI vs.\ 500~PPI), whereas
non-mated pairs are drawn from the 500~PPI pool and are therefore
same-resolution (500~PPI vs.\ 500~PPI).
The two classes consequently differ in a property that has nothing to do
with identity, and a model could in principle separate them by reading
resolution or appearance statistics instead of comparing ridge structure.
We state this plainly at the outset because it governs how the strongest
results in Section~\ref{sec:results} may be read. To settle it we also build a
matched-resolution variant in which the non-mated pairs are formed as
1000~PPI vs.\ 500~PPI, so both classes share the same resolution structure;
Section~\ref{sec:perfect} reports that the strongest result is unchanged under
this control, so the ranking is not an artefact of resolution mismatch.
The collapse results are not affected either, for the
reason given in Section~\ref{sec:limitations}: FAR is computed on non-mated
pairs, which are same-resolution by construction and involve no
cross-resolution comparison at all.
Each pair record also carries demographic metadata (age, gender, and race
of both subjects), which supports the stratified analysis in
Section~\ref{sec:fairness}.

\subsection{Models}

The four open-source models span the systems that lead recent fingerprint and
face benchmarks: \textbf{InternVL3-8B-Instruct}~\cite{internvl3} (bfloat16,
$\sim$16~GB), a top open-source performer in FPBench;
\textbf{Qwen2.5-VL-7B-Instruct}~\cite{qwen25vl}, the architecture strongest in
FaceRecBench; its successor \textbf{Qwen3-VL-8B-Instruct}~\cite{qwen3vl}; and
\textbf{Gemma-3-12B-IT}~\cite{gemma3}, the largest local model. The latter
three use 4-bit NF4 quantization ($\sim$6--8~GB). We add
\textbf{Claude Opus~4.8}~\cite{claudeopus} (Anthropic API, undisclosed weights)
as a high-capability proprietary reference. Exact identifiers are
\texttt{InternVL3-8B-Instruct}, \texttt{Qwen2.5-VL-7B-Instruct},
\texttt{Qwen3-VL-8B-Instruct}, \texttt{Gemma-3-12B-IT}, and
\texttt{claude-opus-4-8}; all runs were collected in June~2026.

The open-source models run via HuggingFace \texttt{transformers} on a single
NVIDIA RTX~4080~SUPER (16.9~GB) with deterministic decoding
(\texttt{temperature}$=0$); Claude is queried as independent stateless
requests. The two images and the text prompt are the only inputs to any
model: no filenames, subject IDs, resolution labels, or demographics appear in
a request, so no model has access to ground-truth identity at inference.

\subsection{Prompting Strategies}
\label{sec:prompts}

We evaluate three prompting strategies, using a fixed system prompt
``\textit{You are an expert fingerprint examiner.}'' across all settings.
Before inference, every image is converted to grayscale, contrast-normalized
(autocontrast), and resized to $448\times448$ pixels, identically for all
five models; this common preprocessing also equalizes the pixel dimensions
of the 500 and 1000~PPI images, though it does not remove the underlying
near-duplicate relationship between a mated pair
(Section~\ref{sec:discussion}).
Under \textbf{zero-shot} the model receives the two images and the binary
question ``Do these two images belong to the same person? (A) Yes (B) No.''
\textbf{Task description} prepends a sentence explaining that repeated captures
of one finger vary slightly while ridge patterns persist, before the same
binary question. \textbf{Similarity scoring} instead asks for a single integer
from 0 (certainly different) to 100 (certainly the same). The three templates
are reproduced verbatim in Fig.~\ref{fig:supp_prompts}.

\subsection{Response Parsing and Metrics}

For binary prompts we parse responses with the three-step fallback of
VLMEvalKit~\cite{vlmevalkit} (leading A/B; then A/B anywhere; then keyword
matching on SAME/YES/GENUINE vs.\ DIFFERENT/NO/IMPOSTOR), marking anything
unmatched INVALID. For the scoring prompt we take the first integer via regex
and clamp it to $[0,100]$.
We report standard verification metrics: for binary prompts, accuracy, FAR,
FRR, and collapse detection (any answer chosen $>$95\% of the time); for
scoring, mean confidence on mated and non-mated pairs, their separation
$\Delta$, EER, and AUC from a threshold sweep over $[0,100]$.

% ────────────────────────────────────────────────────────────────────────────
\section{Results}
\label{sec:results}

\subsection{Binary Prompting: Collapse Under Task Description}

\textbf{Positive-bias collapse} appears in five of ten binary
configurations, with FAR reaching 96.4--100\%
(Table~\ref{tab:main_results}). All four open-source models collapse under
the task-description prompt, and Gemma-3-12B collapses under zero-shot as
well. The proprietary Claude Opus~4.8 is the sole exception, resisting
collapse under both prompts.

% ── Unified results table (binary + similarity scoring) ─────────────────────
\begin{table}[t]
\centering
\caption{SLAPBench verification results on NIST SD302b (7,832 pairs:
  176~genuine $+$ 7,656~impostor). Models are ordered by similarity-scoring
  AUC. \textbf{Binary prompting} reports accuracy and FAR for zero-shot~(ZS)
  and task-description~(TD); red cells mark collapsed configurations
  (FAR~$\geq$95\%, the model answers ``same person'' on nearly every pair),
  the green cell marks the single strongest binary configuration.
  \textbf{Similarity scoring} reports mean confidence on genuine~(Gen) and
  impostor~(Imp) pairs, separation~$\Delta$, EER, AUC (shaded green$\rightarrow$red
  for strong$\rightarrow$random discrimination), and TAR at FAR~$=0.1\%$.
  Low binary accuracy in collapsed runs ($\approx$2--6\%) reflects the
  43:1 impostor-to-genuine imbalance; FAR, AUC, and EER are the meaningful
  measures. Claude Opus~4.8 is the only model that resists collapse under
  \emph{both} prompts and attains the strongest binary configuration
  (ZS FAR 20.2\%).}
\label{tab:main_results}
\renewcommand{\arraystretch}{1.25}
\setlength{\tabcolsep}{4.5pt}
\footnotesize
\resizebox{\textwidth}{!}{%
\begin{tabular}{c l cccc cccccc}
\toprule
& & \multicolumn{4}{c}{\cellcolor{grpBin}\textbf{Binary Prompting}}
  & \multicolumn{6}{c}{\cellcolor{grpScore}\textbf{Similarity Scoring (0--100)}} \\
\cmidrule(lr){3-6}\cmidrule(lr){7-12}
\textbf{\#} & \textbf{Model}
  & \textbf{ZS Acc} & \textbf{ZS FAR} & \textbf{TD Acc} & \textbf{TD FAR}
  & \textbf{Gen} & \textbf{Imp} & \boldmath{$\Delta$} & \textbf{EER}
  & \textbf{AUC} & \textbf{TAR\textsubscript{0.1\%}} \\
\midrule
1 & Qwen3-VL-8B
  & 74.1\% & 26.5\% & 3.3\% & \cellcolor{collapseRed}98.9\%
  & \textbf{100.0} & 63.8 & $+$36.2 & \textbf{0.00\%}
  & \cellcolor{aucPerfect}\textbf{1.000} & \textbf{100.0\%} \\
2 & Claude Opus 4.8
  & \textbf{80.3\%} & \cellcolor{funcGreen}\textbf{20.2\%} & 50.2\% & 50.9\%
  & 91.3 & 48.4 & $+$42.9 & 11.75\%
  & \cellcolor{aucStrong}0.953 & 56.8\% \\
3 & Gemma-3-12B
  & 5.7\% & \cellcolor{collapseRed}96.4\% & 2.3\% & \cellcolor{collapseRed}100.0\%
  & 75.8 & 63.3 & $+$12.5 & 15.10\%
  & \cellcolor{aucMid}0.837 & 23.3\% \\
4 & InternVL3-8B
  & 29.3\% & 72.3\% & 2.2\% & \cellcolor{collapseRed}100.0\%
  & 57.7 & 79.1 & $-$21.4 & 48.09\%
  & \cellcolor{aucWeak}0.590 & 0.0\% \\
5 & Qwen2.5-VL-7B
  & 32.0\% & 69.6\% & 2.2\% & \cellcolor{collapseRed}100.0\%
  & 95.0 & 90.1 & $+$4.9 & 43.34\%
  & \cellcolor{aucBad}0.567 & 0.0\% \\
\bottomrule
\end{tabular}%
}
\end{table}

All five collapsed runs sit at 2.2--5.7\% accuracy rather than 50\%: accepting
all 7,832 pairs scores only the 176 mated pairs (2.25\%) correct, so FAR, at
96.4--100\%, is the meaningful measure of a complete verification failure.

Under zero-shot prompting, four models avoid collapse (FAR 20.2--72.3\%,
Table~\ref{tab:main_results}). \textbf{Claude Opus~4.8 gives the strongest
binary result overall} (80.3\% accuracy, FAR 20.2\%), ahead of the best
open-source model, Qwen3-VL-8B (74.1\%, FAR 26.5\%); InternVL3-8B and
Qwen2.5-VL-7B trail at FAR 72.3\% and 69.6\%. Gemma-3-12B is the exception,
collapsing even under zero-shot (FAR 96.4\%). Across these non-collapsed
runs nearly every mated pair is accepted (FRR~$\leq$0.6\%), so ``different''
answers fall almost exclusively on impostor pairs.

Adding domain context (TD) collapses all four open-source models to
near-100\% FAR, including Qwen3-VL-8B which had the best open-source ZS
result. \textbf{Claude Opus~4.8 is the lone exception}: under TD it remains
balanced (50.2\% accuracy, FAR 50.9\%, answering ``same person'' on only
52\% of pairs), neither collapsing nor degrading to the open-source
failure mode.
For the open-source models, prompting complexity increases the false
accept rate rather than reducing it, confirming the FaceXBench and FPBench
observation that additional context does not help and actively hurts in
biometric binary tasks~\cite{facexbench2024,fpbench2025}.

\subsection{Similarity Scoring: Discriminative Signal Revealed}
\label{sec:scoring}

\begin{figure}[tbp]
  \centering
  \includegraphics[width=\linewidth]{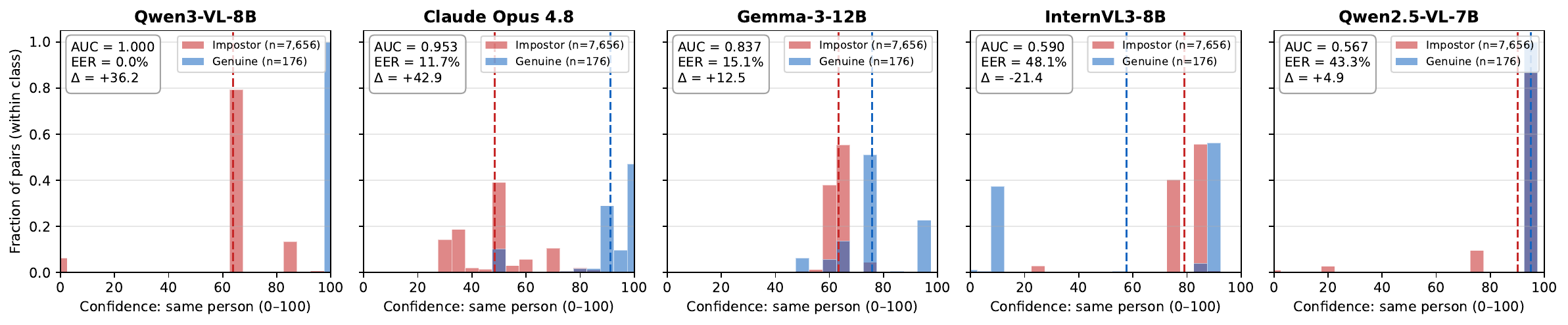}
  \caption{Score distributions under the similarity-scoring prompt
    (7,832 pairs). Bars give the fraction of pairs \emph{within each class},
    normalized separately to offset the 176:7{,}656 mated-to-non-mated
    imbalance; dashed lines mark per-class means; each panel is annotated with
    AUC, EER and mean separation $\Delta$. Scores are discrete, so no
    continuous density is fitted. The first three models discriminate;
    InternVL3-8B is inverted ($\Delta = -21.4$, with a bimodal mated
    distribution) and Qwen2.5-VL-7B is compressed, piling both classes
    near 95.}
  \label{fig:score_dist}
\end{figure}

The similarity-scoring prompt eliminates collapse across all four
open-source models: no model produces a single constant output, and mated
and non-mated pairs occupy distinguishable (though not always separable)
score ranges. Figure~\ref{fig:score_dist} shows the distributions for all
five models, and the scoring block of Table~\ref{tab:main_results} gives the
metrics.

The models split into two tiers (Table~\ref{tab:main_results}).
\textbf{Qwen3-VL-8B} separates the classes completely (AUC~$= 1.000$,
EER~$= 0.0\%$); a result this clean calls for scrutiny rather than
celebration, and we examine it in Section~\ref{sec:perfect}.
\textbf{Claude Opus~4.8} is the strongest graded discriminator
(AUC~$= 0.953$), with the widest mated-to-non-mated gap of any model
($+$42.9~points), and \textbf{Gemma-3-12B} is functional but weaker
(AUC~$= 0.837$).
The other two fail in opposite ways: \textbf{InternVL3-8B} is
\emph{inverted}, reporting higher confidence on impostors than on genuine
matches ($\Delta = -21.4$, AUC~$= 0.590$), while \textbf{Qwen2.5-VL-7B} is
\emph{compressed}, over 90\% confident on nearly every pair regardless of
identity ($\Delta = +4.9$, AUC~$= 0.567$). Both are near chance and neither
admits a threshold that accepts a mated pair before an impostor, so both
reach TAR~$= 0.0\%$ at FAR~$= 0.1\%$. Per-hand behaviour is symmetric for
every model, so we omit the FRGP~13/14 split.

Scoring therefore eliminates collapse in all four open-source models without
guaranteeing meaningful discrimination: Qwen3-VL-8B, Claude, and Gemma
separate the classes, whereas InternVL3-8B and Qwen2.5-VL-7B stay effectively
random.
\textbf{The binary collapse in Table~\ref{tab:main_results} is thus a
prompt-format artifact; whether discriminative capability can be
unlocked by scoring depends on model architecture.}
Two real pairs in Fig.~\ref{fig:supp_qualitative} illustrate this: on a
mated pair only InternVL3-8B fails (returning~9), and on a non-mated pair only
InternVL3-8B~(85) and Qwen2.5-VL-7B~(95) falsely accept.

\begin{figure}[tbp]
  \centering
  \includegraphics[width=0.48\linewidth]{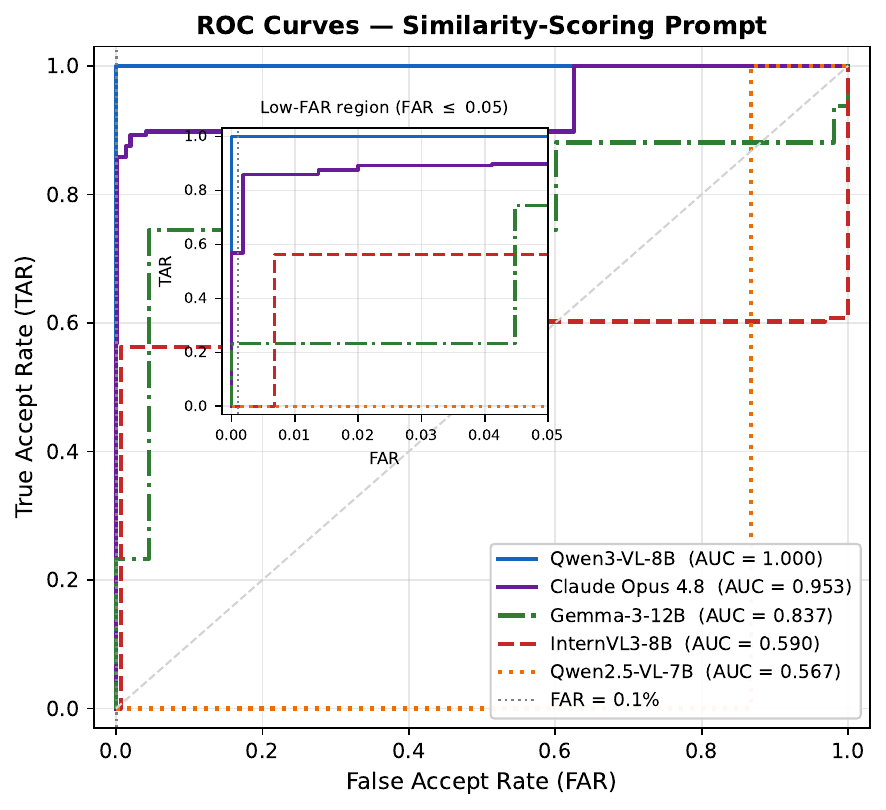}
  \caption{ROC curves under the similarity-scoring prompt (7,832 pairs)
    for the four open-source models and the proprietary Claude Opus~4.8.
    The inset magnifies the low-FAR region (FAR $\leq 0.05$). The dotted
    vertical line marks FAR~$= 0.1\%$.
    Qwen3-VL-8B (AUC~$= 1.000$) achieves perfect discrimination, Claude
    Opus~4.8 (AUC~$= 0.953$) and Gemma-3-12B (AUC~$= 0.837$) are clearly
    functional, while InternVL3-8B and Qwen2.5-VL-7B perform near random
    (AUC~$= 0.590$ and $0.567$).}
  \label{fig:roc}
\end{figure}

Figure~\ref{fig:roc} plots the full ROC curves for the four open-source
models and Claude Opus~4.8.
The inset shows the low-FAR operating region (FAR $\leq 0.05$),
where Qwen3-VL-8B achieves TAR~$= 100\%$ even at FAR~$= 0.1\%$.
The TAR\textsubscript{0.1\%} column of Table~\ref{tab:main_results}
reports TAR at FAR~$= 0.1\%$, the standard operational stringency target in
border biometrics. InternVL3-8B and Qwen2.5-VL-7B reach 0\% here because
their score distributions offer no threshold that holds FAR below 0.1\%
while still accepting any mated pair, a direct consequence of inverted
calibration (InternVL3) and compressed scoring (Qwen2.5).

\subsection{Demographic Fairness}
\label{sec:fairness}

FaceRecBench reports that the best face-verification MLLM also carries the
strongest demographic bias~\cite{facerecbench2025}; we test the pattern on
SLAP by recomputing similarity-scoring performance within gender, race, and
age subgroups for the three discriminating models (Fig.~\ref{fig:fairness};
full protocol and per-subgroup table in Appendix~\ref{app:fairness}). The pattern
runs
\emph{opposite} to FaceRecBench's: on SLAP data the stronger models look
fairer. Claude Opus~4.8 varies by at most 1.5~AUC points across gender, whereas
the weakest discriminator, Gemma-3-12B, shows the widest spread; Qwen3-VL-8B is
uniformly perfect, equally consistent with a group-invariant representation and
with the non-identity shortcut of Section~\ref{sec:perfect}. To the extent the
small subgroups support any reading, disparity tracks model \emph{weakness}
rather than strength. We treat this as an initial, hypothesis-generating probe
on a single database rather than an audit.

\begin{figure}[tbp]
  \centering
  \includegraphics[width=\linewidth]{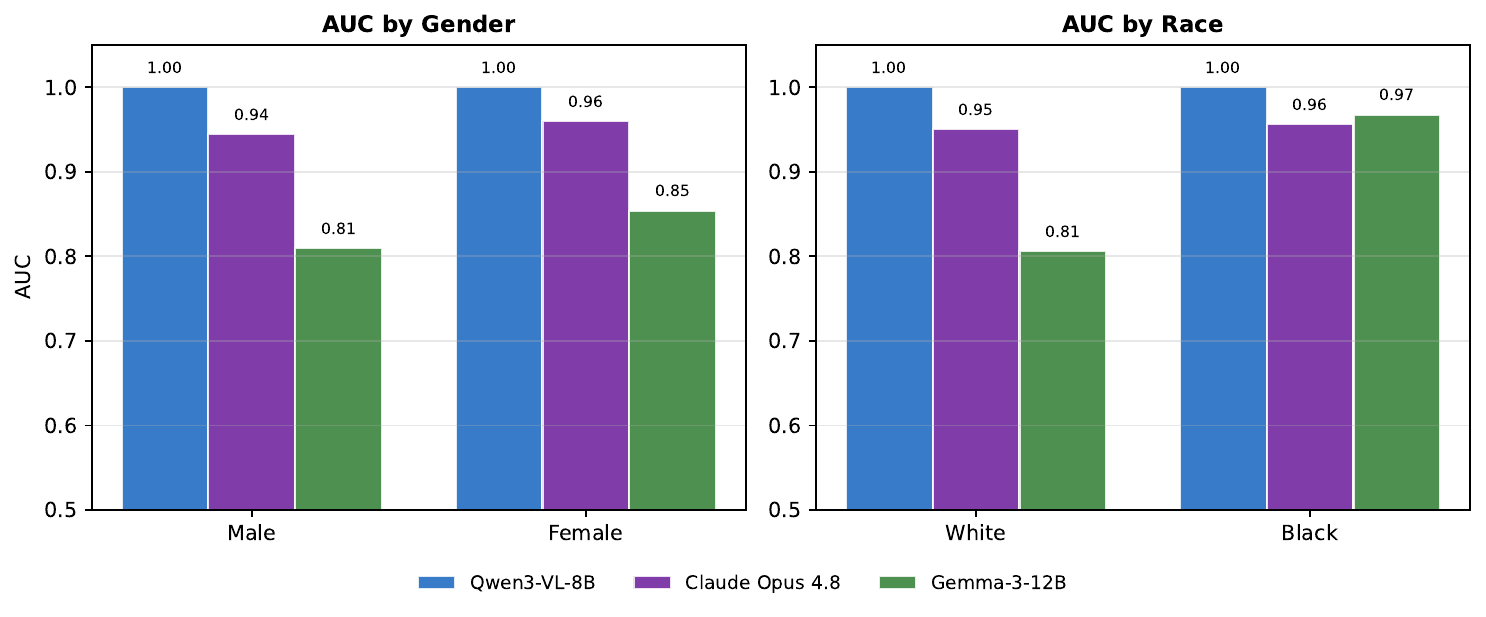}
  \caption{Per-subgroup AUC by gender and race for the three discriminating
    models (age bands and exact values in Table~\ref{tab:supp_fairness}).
    Qwen3-VL-8B is uniformly perfect; Claude Opus~4.8 varies by
    $\leq$1.5~AUC points across gender; Gemma-3-12B shows the largest
    disparities. The Black group is small ($n_i = 272$), so Gemma's high
    value there should be read with caution.}
  \label{fig:fairness}
\end{figure}

% ── Removed per advisor request: "Relationship to Prior Work" subsection. ──
% Kept commented here so the arXiv body matches the reviewed version exactly.
% \subsection{Relationship to Prior Work}
%
% We do not compare numbers directly with FPBench~\cite{fpbench2025}: it uses
% individual rolled and plain impressions under a four-option MCQ format,
% whereas SLAPBench uses four-finger captures under binary or continuous
% scoring, so the datasets, image types, and metrics are not comparable. What
% both establish, and ours confirms, is the qualitative pattern that binary
% prompting on fingerprints collapses open-source models regardless of image
% type. The two benchmarks are complementary rather than competing.

% ────────────────────────────────────────────────────────────────────────────
\section{Discussion}
\label{sec:discussion}

\subsection{Why Binary Prompts Collapse on SLAP Data}

Collapse follows a consistent signature: every collapsed run has FRR~$= 0\%$
and FAR~$\geq 96.4\%$, the mark of a model whose prior for ``same person'' is
so strong that no visual evidence moves it toward ``B.'' Two structural
factors make this prior stronger for SLAP than for the single-finger images
of FPBench. First, any two SLAP images share the same gross morphology (four
fingers, similar spacing, skin texture, and orientation), so their global
similarity is far higher than between two isolated rolled impressions,
whether the pair is mated or not; the surface-level similarity prior is thus
satisfied for every pair regardless of identity. Second, the task-description
prompt, which states that ``the same person's fingers produce slightly
different images each time,'' \emph{increases} collapse: it primes a schema in
which variation is a property of mated pairs and lowers the ``same'' threshold
across all inputs, the model having read ``same person'' twice before seeing
either image. The lesson for benchmark design is that naturalistic domain
descriptions can worsen collapse by legitimizing visual differences.

This also explains the exception. Claude Opus~4.8 resists collapse under both
prompts exactly as the mechanism predicts, rejecting about half of impostors
even under task description (FAR 50.9\%) and four-fifths under zero-shot
(FAR 20.2\%). A model with identity evidence strong enough to overcome the
similarity prior is not forced into the degenerate ``same'' response. Collapse
is therefore a capability threshold, not a fixed property of the binary
format: the same format that collapses four open-source models leaves the
strongest one balanced.

\subsection{Why Similarity Scoring Breaks Collapse}

The scoring prompt resolves the same underlying problem by changing what
the model is asked to optimize.
Under binary prompting, the model must commit to a categorical label
against its own prior; the prior wins.
Under scoring, the model places a continuous estimate on a calibrated
scale, a task that requires reporting a degree of confidence rather than
crossing a decision boundary.
This framing lets the model's uncertainty about impostor pairs surface as a
lower score instead of being suppressed into the same ``A'' response it
gives to mated pairs.

The scoring prompt eliminates collapse in all four open-source models, but
the resulting distributions show that scoring is necessary without being
sufficient. It reveals whether discriminative capability exists; it does
not confer it. The two models that lack it are InternVL3-8B and
Qwen2.5-VL-7B (Fig.~\ref{fig:score_dist}, rightmost panels).

The two failures differ in kind. \textbf{InternVL3-8B} is \emph{inverted},
scoring impostors above genuine pairs; we suspect its encoder keys on
fine-grained ridge texture, which contrasts more strongly across two
different subjects than between two renderings of one capture, so scoring
faithfully reports this signal in the wrong direction. \textbf{Qwen2.5-VL-7B}
is instead \emph{compressed}: correctly ordered but confined to a narrow
high-confidence band, the scoring-domain echo of the same similarity prior
that drives its binary collapse. In neither case can scoring manufacture
discrimination the representation does not encode.

Why does scoring unlock the stronger models but not these? The most
actionable explanation is that its calibrated anchors tell the model
\emph{where to look} (ridge flow, minutiae) and \emph{what to report}
(a value on a fixed scale) rather than forcing a binary commit, and only
models with sufficient visual reasoning (Qwen3, Claude) can act on them.
The practical implication is that \textbf{a calibrated scoring prompt is a
more effective probe of MLLM biometric capability than binary forced choice,
but only when the architecture can exploit it.}

\subsection{A Perfect Score Is a Diagnostic, Not a Capability Claim}
\label{sec:perfect}

Qwen3-VL-8B attains AUC~$= 1.000$ with EER~$= 0.0\%$. We do not read this as
a capability result: the number is explained by the structure of the protocol
rather than by verification ability, and the episode is the paper's most useful
finding for benchmark designers, because it shows how readily a SLAP protocol
can manufacture a perfect score.

The behavior underlying the number is revealing. Qwen3-VL-8B assigns
\emph{exactly} 100 to all 176 mated pairs, while its non-mated scores take only
seven distinct values ($\{0, 30, 45, 65, 85, 90, 95\}$, mean 63.8) and never
exceed 95. Perfect separation is therefore not a finely graded judgment but a
coarsely quantized output in which the single response 100 is reserved for the
mated class; any mechanism that saturates the mated class this way reproduces
AUC~$= 1.000$ whether or not it compares ridge structure. A model doing genuine
forensic comparison would show some graded uncertainty across 176 subjects, as
Claude Opus~4.8 does (nine distinct mated and 48 non-mated values on the same
pairs).

The \textbf{resolution shortcut} is the most economical explanation and the one
we set out to test: mated pairs are cross-resolution (1000 vs.\ 500~PPI) while
non-mated pairs are same-resolution (500 vs.\ 500), so a model that merely
detects downsampling could separate the classes without comparing identity. We
therefore ran the \textbf{matched-resolution control} of
Section~\ref{sec:pairs}, rebuilding every non-mated pair as 1000~PPI vs.\
500~PPI so both classes share the same structure and differ only in identity
(FRGP~13: 88 mated, all 3{,}828 non-mated). The result is essentially
unchanged: exactly 100 on every mated pair, no non-mated pair above 95,
AUC~$= 1.000$ with the non-mated mean moving only from 65.6 to 65.3. Making
resolution non-diagnostic had no measurable effect, so \textbf{we exclude the
resolution shortcut}.

What the control cannot remove is \textbf{near-duplicate detection}: a mated
pair is one capture rendered at two scales, so the two images share far
stronger low-level correspondence than a deployed matcher sees between
independent impressions, and rearranging resolutions cannot disturb this.
SD302b cannot adjudicate it, because Device~R holds one capture per subject and
finger and the Device~R and Device~S pools are disjoint, so no mated pair can
be formed from two independent impressions; settling the question needs a
database with repeat SLAP captures. \textbf{Memorization} we set aside: had the
model memorized identity labels the knowledge should transfer across prompts,
yet its zero-shot FAR is 26.5\% and it collapses under task description, and no
other model (not even Claude at AUC~$= 0.953$) reproduces the result.

The honest summary is narrow: Qwen3-VL-8B separates SLAPBench's pairs perfectly
under scoring; the resolution shortcut has been tested and ruled out;
near-duplicate detection remains and cannot be excluded within SD302b; and
genuine identity discrimination, though consistent with the control, cannot be
separated from it here, so no operational verification capability should be
inferred. Near-perfect MLLM biometric results warrant a protocol audit before
they warrant belief.

\subsection{The Left-Hand / Right-Hand Asymmetry}

A modest per-hand asymmetry appears under zero-shot prompting, but its
direction is model-specific: Qwen3-VL-8B favours the left hand by 5.0~FAR
points while Claude Opus~4.8 favours the right by 4.7, and the other three
models are near-symmetric (FAR gaps $\leq$3.6~points). Since both FRGPs draw
identical pair counts from the same 88 subjects, this reflects model-specific
pretraining priors rather than any property of SLAP images; disentangling it
would need a second dataset.

\subsection{Limitations}
\label{sec:limitations}

Three limitations bound what the present results establish, and we state
them alongside the findings each one does and does not touch.

\textbf{Mated-pair construction.}
SD302b provides a single SLAP capture per subject and finger position, so a
mated pair can only be formed across the two resolutions of that one capture
(Section~\ref{sec:dataset}). Our matched-resolution control shows this
asymmetry does not drive the results (Section~\ref{sec:perfect}); what the
single-capture design leaves in place is that a mated pair is one exposure
rendered twice, so near-duplicate detection cannot be separated from identity
matching within this dataset, and settling it requires a database with repeat
captures. This is why we decline to read Qwen3-VL-8B's AUC~$= 1.000$ as
verification capability. It does not bear on the central findings: collapse is
measured on non-mated pairs alone, so the collapse of the open-source models,
Claude Opus~4.8's resistance, and the removal of collapse by scoring are all
unaffected.

\textbf{Model coverage.}
Our proprietary coverage is a single system. FPBench reports that Gemini
2.5 Pro leads open-source models by roughly 4 percentage points on
single-finger tasks~\cite{fpbench2025}, and whether the collapse resistance
we observe in Claude Opus~4.8 is a property of frontier-scale models
generally or specific to Claude cannot be settled with one data point.

\textbf{Scope.} The fairness analysis (Section~\ref{sec:fairness}) rests on
88~subjects with several small subgroups, so it is an initial probe rather than
an audit; and SLAPBench compares MLLMs against one another, not against a
deployed matcher such as VeriFinger~\cite{verifinger}, so it places models on a
relative rather than an absolute scale. Both are left to future work.

\subsection{Future Work}

Several directions follow. The most immediate is the
\textbf{matched-resolution control} of Section~\ref{sec:perfect}, rebuilding
the non-mated pairs as 1000-vs-500 comparisons (in both orderings) so that
resolution no longer signals class, alongside a mated set from a database with
repeat impressions to remove the near-duplicate cue. Beyond that:
\textbf{broader proprietary
evaluation} (GPT-series, Gemini) to test whether Claude's collapse resistance
is general; \textbf{domain-adaptive fine-tuning}, which FPBench shows adds
7--39\% on single-finger tasks~\cite{fpbench2025}; \textbf{larger-scale
fairness} on a balanced multi-database pool; \textbf{algorithmic-baseline
calibration} against a matcher such as VeriFinger; and \textbf{benchmark
extension} to the seven further SLAPBench tasks (finger counting, laterality,
localization, position, quality, rotation, and segmentation-challenge
identification).

% ────────────────────────────────────────────────────────────────────────────
\section{Conclusion}
\label{sec:conclusion}

We introduced SLAPBench, the first benchmark for evaluating multimodal
large language models on four-finger SLAP fingerprint verification, built
on NIST SD302b with 7,832 pairs (176 mated, 7,656 non-mated).
Task-description prompting collapses all four open-source models to
near-100\% FAR and Gemma-3-12B collapses under zero-shot as well, while
Claude Opus~4.8 resists collapse under both prompts and records the
strongest binary result (FAR~$= 20.2\%$).
Similarity scoring eliminates collapse across the open-source models and
shows that architecture sets the ceiling, with discrimination ranging from
functional (Claude Opus~4.8, Gemma-3-12B) to near-random or inverted
(Qwen2.5-VL-7B, InternVL3-8B; full metrics in Table~\ref{tab:main_results}).
Collapse is thus an artifact of the prompt format, whereas unlocking
discriminative capability through scoring requires sufficient visual
reasoning.
Qwen3-VL-8B's perfect separation (AUC~$= 1.000$) we deliberately do not claim
as capability: a matched-resolution control rules out the resolution
shortcut, but SD302b's single-capture design still admits near-duplicate
detection, so whenever a dataset offers only one capture per subject the
synthesized mated class can itself become the cue a model reads, and
near-perfect results call for a protocol audit before belief.
With the first SLAP-specific baseline established, domain-adaptive
fine-tuning is the natural next step, consistent with the gains FPBench
reports on single-finger tasks~\cite{fpbench2025}.

% ────────────────────────────────────────────────────────────────────────────
\bibliographystyle{splncs04}
\bibliography{references}

% ════════════════════════════════════════════════════════════════════════════
%  APPENDIX
%  This was supplementary.tex in the ECCV submission, where it had to be a
%  separate PDF. arXiv has no page limit, so it is merged here in full.
%  Content is unchanged apart from cross-references, which now point at the
%  real sections/figures via \ref instead of hard-coded numbers.
% ════════════════════════════════════════════════════════════════════════════
\clearpage
\appendix

% These pages carry large pinned floats and cannot fill to the bottom margin.
% Under the class's default flush-bottom, TeX absorbs the slack by stretching
% the glue after a heading, which opens a ~160pt hole between the section title
% and its first paragraph. \raggedbottom lets the slack fall at the foot of the
% page instead, where it belongs.
\raggedbottom

% Number appendix floats S1, S2, ... so they stay distinct from the main
% paper's Figs. 1-3 and Table 1.
\setcounter{figure}{0}
\setcounter{table}{0}
\renewcommand{\thefigure}{S\arabic{figure}}
\renewcommand{\thetable}{S\arabic{table}}

\section{Supporting Figures}
\label{app:figures}

This appendix collects the supporting figures referenced in the main text,
together with the full protocol for the fairness analysis. It introduces no
new claims; every result it illustrates is stated and quantified in the main
text.

\begin{figure}[H]
  \centering
  \includegraphics[width=0.55\linewidth]{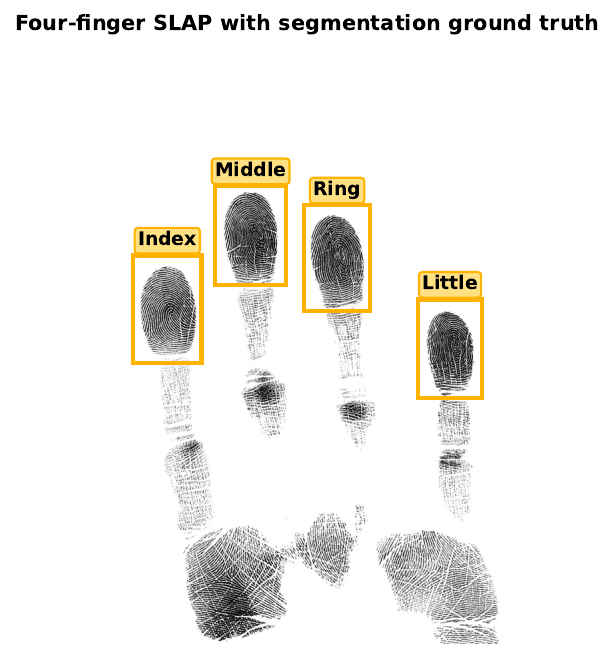}
  \caption{\textbf{Four-finger SLAP capture with segmentation ground truth}
    (NIST SD302b, Device~R, 500~PPI; referenced in
    Section~\ref{sec:dataset}). The per-finger regions (index, middle, ring,
    little) come from the segmentation coordinates distributed with SD302b. A
    single frame holds four overlapping fingers of broadly similar morphology,
    so any two SLAP images look globally alike whether they are mated or not,
    which underlies the positive-bias collapse analyzed in
    Section~\ref{sec:discussion}.}
  \label{fig:supp_anatomy}
\end{figure}

\begin{figure}[H]
  \centering
  \includegraphics[width=0.9\linewidth]{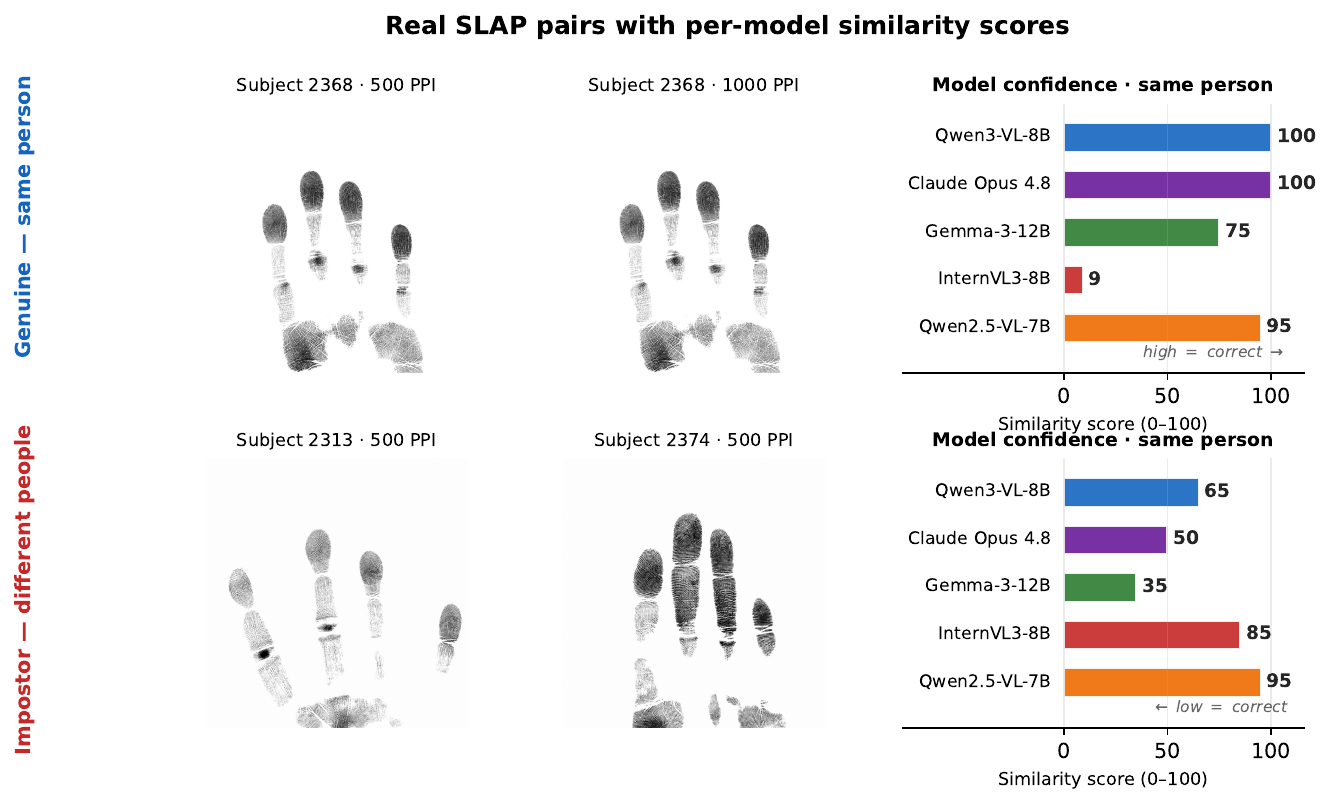}
  \caption{\textbf{Real SLAP pairs with each model's similarity score
    (0--100)}, referenced in Section~\ref{sec:scoring}.
    \textbf{Top:} a mated pair (subject~2368, 500 vs.\ 1000~PPI); every model
    but InternVL3-8B reports high confidence, while InternVL3 inverts and
    returns~9. \textbf{Bottom:} a non-mated pair (subjects~2313 and~2374);
    Qwen3-VL-8B, Claude, and Gemma assign lower confidence, whereas
    InternVL3-8B~(85, inverted) and Qwen2.5-VL-7B~(95, saturated) falsely
    accept it. The italic cue marks which direction is correct for each pair.}
  \label{fig:supp_qualitative}
\end{figure}

\begin{figure}[H]
\centering
\footnotesize
\setlength{\fboxsep}{6pt}
\noindent\fbox{\parbox{0.95\linewidth}{%
\textbf{System prompt (all strategies).}\enspace
{\ttfamily You are an expert fingerprint examiner.}
\vspace{4pt}\hrule\vspace{4pt}
\textbf{Zero-shot (ZS).}\\
{\ttfamily Below are two SLAP fingerprint images. A SLAP image captures four
fingers from one hand pressed simultaneously on a scanner.\\
Do these two images belong to the same person?\\
(A) Yes, same person\quad(B) No, different people\\
Reply with only the letter A or B.}
\vspace{4pt}\hrule\vspace{4pt}
\textbf{Task description (TD).}\\
{\ttfamily A SLAP fingerprint image captures four fingers from one hand
simultaneously. The same person's fingers produce slightly different images
each time they press the scanner (different pressure, slight movement), but
the ridge patterns remain the same. Below are two SLAP fingerprint images. Do
these two images belong to the same person?\\
(A) Yes, same person\quad(B) No, different people\\
Reply with only the letter A or B.}
\vspace{4pt}\hrule\vspace{4pt}
\textbf{Similarity scoring (SS).}\\
{\ttfamily You are a forensic fingerprint examiner analyzing two SLAP
fingerprint images. A SLAP image captures four fingers (index, middle, ring,
little) pressed simultaneously on a scanner. Your task: Examine the ridge flow
patterns, ridge endings, bifurcations, and the relative shape and spacing of
each finger across both images. Then return a single integer between 0 and 100
representing your confidence that both images belong to the same person, where
0 = certain they are different people, 50 = completely uncertain, and
100 = certain they are the same person. A score of 65 means you are 65\%
confident the images are from the same person. Reply with the number only. Do
not include any text, explanation, or label.}
}}
\caption{The three prompt templates, reproduced verbatim (referenced in
  Section~\ref{sec:prompts}). Every model receives the same system prompt
  and, per strategy, the same user prompt, with the two SLAP images attached to
  the user message. \texttt{\%} denotes a literal percent sign.}
\label{fig:supp_prompts}
\end{figure}

\section{Demographic Fairness Detail}
\label{app:fairness}

This section gives the full protocol and per-subgroup table for the fairness
analysis summarized in Section~\ref{sec:fairness}. We recompute
similarity-scoring performance within demographic subgroups: a mated pair is
assigned to a subgroup by its subject's attribute, and a non-mated pair is
counted within a subgroup only when \emph{both} subjects share that attribute.
The analysis is restricted to the three discriminating models (subgroup
breakdowns of the two near-random models are uninformative) and to subgroups
with sufficient support among the 88~clean subjects (gender; White and Black
race; three age bands); Asian, ``other,'' and ``no answer'' race groups
($\leq 6$ subjects) are too small to evaluate reliably.

\begin{table}[H]
\centering
\caption{Per-subgroup similarity-scoring performance (AUC / EER\%) for the
  three discriminating models. $n_g$/$n_i$ are the mated and within-group
  non-mated pair counts. Small subgroups (Black, 18--30) are reported with
  caution. Qwen3-VL-8B is uniform at AUC~$= 1.000$ across all groups, which
  must be read alongside the perfect-score caveat of
  Section~\ref{sec:perfect}.}
\label{tab:supp_fairness}
\renewcommand{\arraystretch}{1.2}
\setlength{\tabcolsep}{5pt}
\begin{tabular}{l rr ccc}
\toprule
\textbf{Subgroup} & \textbf{$n_g$} & \textbf{$n_i$}
  & \textbf{Qwen3} & \textbf{Claude} & \textbf{Gemma} \\
\midrule
Male    &  70 & 1190 & 1.000 / 0.0 & 0.945 / 13.2 & 0.810 / 15.6 \\
Female  & 106 & 2756 & 1.000 / 0.0 & 0.960 / 10.5 & 0.854 / 14.8 \\
\midrule
White   & 122 & 3660 & 1.000 / 0.0 & 0.950 /  7.4 & 0.806 / 18.3 \\
Black   &  34 &  272 & 1.000 / 0.0 & 0.957 / 11.6 & 0.968 /  4.4 \\
\midrule
18--30  &  26 &  156 & 1.000 / 0.0 & 0.967 /  6.1 & 0.815 / 19.2 \\
31--45  &  76 & 1406 & 1.000 / 0.0 & 0.944 / 13.8 & 0.826 / 15.8 \\
46+     &  74 & 1332 & 1.000 / 0.0 & 0.957 / 10.6 & 0.858 / 13.5 \\
\bottomrule
\end{tabular}
\end{table}

Claude Opus~4.8 varies by at most 1.5~AUC points across gender (male 0.945,
female 0.960) and 0.7 across White/Black. Gemma-3-12B shows the widest spread,
with a 4.4-point gender gap and an EER swing from 18.3\% to 4.4\% between White
and Black; that Black subgroup rests on only 272 within-group non-mated pairs
from 17 subjects, so neither its values nor any disparity from them should be
over-read.

\end{document}